\pdfoutput=1
\documentclass{article}
\PassOptionsToPackage{numbers,compress}{natbib}

\usepackage[preprint]{neurips_2026}

\usepackage[utf8]{inputenc}
\usepackage[T1]{fontenc}
\usepackage{hyperref}
\usepackage{url}
\usepackage{booktabs}
\usepackage{amsmath}
\usepackage{amssymb}
\usepackage{amsfonts}
\usepackage{nicefrac}
\usepackage{microtype}
\usepackage{xcolor}
\usepackage{pifont}
\usepackage{tikz}
\usetikzlibrary{positioning,arrows.meta,shapes.geometric,fit,backgrounds,
                 decorations.pathreplacing,calc}
\usepackage{pgfplots}
\pgfplotsset{compat=1.18}
\usepackage{capt-of}

\newsavebox{\tikzfitbox}
\let\origtikzpicture\tikzpicture
\let\origendtikzpicture\endtikzpicture
\renewenvironment{tikzpicture}[1][]{%
  \begin{lrbox}{\tikzfitbox}%
  \origtikzpicture[#1]%
}{%
  \origendtikzpicture%
  \end{lrbox}%
  \ifdim\wd\tikzfitbox>\linewidth
    \resizebox{\linewidth}{!}{\usebox{\tikzfitbox}}%
  \else
    \usebox{\tikzfitbox}%
  \fi
}
\usepackage{array}
\usepackage{multirow}
\usepackage{enumitem}

\newcommand{\R}{\mathbb{R}}
\newcommand{\softmax}{\operatorname{softmax}}
\newcommand{\mem}{\mathrm{Mem}}
\newcommand{\freeze}{\mathrm{frozen}}
\newcommand{\topk}{\operatorname{top\text{-}k}}

\title{Trained Persistent Memory for Frozen Decoder-Only LLMs}

\author{%
  Hong Jeong \\
  Inha University in Tashkent, Uzbekistan \\
  \texttt{hjeong@postech.ac.kr}
}

\begin{document}
\maketitle

\begin{abstract}
Decoder-only language models are stateless: hidden representations are
discarded after every forward pass and nothing persists across sessions.
Jeong (2026a) showed that trained memory adapters give a frozen
encoder--decoder backbone persistent latent-space memory, building on
the lateral-memory framework of Jeong (2026b,\,c).
Here we ask whether the same principle transfers to the decoder-only
setting, where no cross-attention pathway exists and memory must enter
through self-attention alone.
We adapt six methods---prefix, parallel cross-attention, KV extension,
Hebbian memory, context-gated branch, and slot-based sparse write---to a
frozen GPT-2, training only a small adapter~$\theta_{\mem}$.  The write
rule is shared; only the read injection changes from decoder
cross-attention to self-attention KV prefix or parallel branch.
On LoCoMo we find a striking \emph{inductive-bias dichotomy}: at
$1\times$ capacity, three methods with strong architectural
priors---cross-attention~(M.2), Hebbian~(M.4), and slot
write~(M.6)---achieve retained-memory scores of 7--18\% and knowledge
gains $\Delta K$ of 7--10, while the other three fail ($<$0.4\%).
At $10\times$ capacity all six converge, showing the gap is
architectural, not fundamental.  Together with the encoder--decoder
results of Jeong (2026a) and the brain-inspired modules of
Jeong (2026b,\,c), these findings establish persistent latent-space
memory as a general paradigm spanning major transformer families.
\end{abstract}

\section{Introduction}
\label{sec:intro}

Modern decoder-only language models---GPT-2~\citep{radford2019gpt2},
LLaMA~\citep{touvron2023llama}, Phi-2~\citep{javaheripi2023phi2},
Mistral~\citep{jiang2023mistral}---process input autoregressively through
a stack of causal self-attention layers.  The forward pass is:
\begin{equation}\label{eq:stateless}
  H_t = D_{\freeze}(x_t), \qquad \hat{y}_t = \mathrm{head}(H_t),
\end{equation}
where $D_{\freeze}$ is the frozen transformer body, $H_t \in \R^{n \times d}$
is the final hidden state, and $\hat{y}_t$ is the output distribution.
After generation, $H_t$ is discarded.  There is no mechanism for
information to survive from one session to the next.

\citet{jeong2026frozen} introduced persistent latent-space memory
for frozen encoder--decoder models, showing that a small trained adapter
$\theta_{\mem}$ can maintain a memory bank
$P_t \in \R^{n_P \times d}$ that accumulates across turns and enables
\emph{conversational learning}: facts stated in session~1 can be recalled
in session~10 without re-stating them and without million-token context
windows.  The key insight was that the write rule is a differentiable
operation on dense vectors, not a text-level retrieval step, and that
training is necessary for the frozen backbone to discriminate useful
memory entries from noise.

The encoder--decoder setting, however, offered a natural injection point:
the decoder's cross-attention heads, pre-trained to attend to external
encoder outputs, could be re-purposed to attend to memory entries with
minimal architectural disruption.  Decoder-only models lack this pathway.
Their only attention mechanism is causal self-attention, which was
pre-trained to attend to preceding tokens within the \emph{same} sequence.
Injecting persistent memory therefore requires a fundamentally different
read strategy.

We hypothesise, however, that persistent memory is most
naturally aligned with decoder-only architectures, despite the injection
challenge.  The key observation is that autoregressive generation is
inherently \emph{sequential}: each token step already reads from a KV
cache that summarises all preceding context, and persistent memory is
simply a generalisation of this cache to information that survives across
sessions.  Encoder--decoder models, by contrast, process context
bidirectionally in the encoder and must bridge an architectural gap
between the encoder representation and the decoder's memory needs.
Encoder-only models (BERT-style) lack generation entirely and process
input in a single bidirectional pass with no temporal flow---the least
natural setting for incremental memory accumulation.  This motivates
a conjectured ordering of architectural affinity:
\[
  \text{Decoder-only} \;>\; \text{Encoder--decoder} \;>\; \text{Encoder-only}
\]
for persistent latent-space memory.  \citet{jeong2026frozen} already established the
middle case; the present paper tests the strongest case.

This paper makes three contributions:

\begin{enumerate}[leftmargin=*,itemsep=2pt]
\item \textbf{Architecture transfer.}
  We adapt all six memory methods from~\citet{jeong2026frozen} to a
  frozen decoder-only backbone (GPT-2), replacing cross-attention read
  paths with self-attention-compatible alternatives: KV prefix injection,
  parallel cross-attention insertion (Flamingo-style), and gated additive
  branches.

\item \textbf{Inductive-bias dichotomy.}
  At standard capacity ($1\times$), only three methods with strong
  architectural priors succeed (M.2~XAttn, M.4~Hebbian, M.6~Slot),
  while the other three fail.  At $10\times$ capacity, all six methods
  converge, revealing that architectural bias determines efficiency under
  constrained capacity.

\item \textbf{Two-metric evaluation.}
  We evaluate persistent memory on both the absolute forgetting-curve
  protocol (retained-memory score) and a knowledge-accumulation metric
  ($\Delta K$), providing complementary views of memory retention and
  knowledge growth over 30~sessions.
\end{enumerate}

\section{Related Work}
\label{sec:related}

\paragraph{Persistent memory for LLMs.}
Application-level memory systems such as MemGPT~\citep{packer2024memgpt}
and MemoryBank~\citep{zhong2024memorybank} operate at the text level:
facts are stored as natural-language strings and retrieved via search.
These systems are architecture-agnostic by construction but add
inference-time overhead proportional to memory size and cannot benefit
from gradient-based training of the memory pathway.
Our approach operates at the \emph{latent} level: the memory bank
$P \in \R^{n_P \times d}$ stores continuous representations, and both
read and write are differentiable operations inside the forward pass.

\paragraph{KV caching and context extension.}
Decoder-only models routinely cache key--value pairs from earlier tokens
to avoid recomputation~\citep{pope2023efficiently}.  Several works extend
this cache across sequences: Memorizing
Transformers~\citep{wu2022memorizing} retrieve from a growing external
KV store, and $\infty$-former~\citep{martins2022inftyformer} compresses
past KV pairs into a fixed-size summary.  These methods cache raw
activations; ours learns a compressed, trained representation that the
frozen backbone can discriminate.  The distinction is precisely what
\citet{jeong2026frozen} established: without trained projections, accumulated cache
states dilute attention rather than sharpen retrieval.

\paragraph{Recurrent and compressive context extension.}
Transformer-XL~\citep{dai2019transformerxl} introduces segment-level
recurrence that carries hidden states across fixed-length segments,
extending the effective receptive field without recomputation.
The Compressive Transformer~\citep{rae2020compressive} adds a second
level of memory by compressing older activations into a lossy summary,
trading exact recall for longer range.
The Recurrent Memory Transformer~\citep{bulatov2023rmt} appends learned
memory tokens that flow between segments.  These methods extend context
\emph{within} a single document or session; our framework extends memory
\emph{across} sessions, storing persistent representations that
accumulate over an arbitrarily long dialogue history.

\paragraph{Learned external memory.}
Neural Turing Machines~\citep{graves2014ntm} and the Differentiable
Neural Computer~\citep{graves2016dnc} introduced differentiable
read/write access to an external memory matrix, trained end-to-end.
Our memory bank~$P$ shares this differentiable read/write design, but
differs in two respects: the transformer backbone is frozen and only the
memory adapter is trained, and the memory persists across independent
inference sessions rather than within a single forward pass.

\paragraph{Parameter-efficient adaptation.}
Prefix tuning~\citep{li2021prefix} prepends learnable soft tokens to
each layer; LoRA~\citep{hu2022lora} injects low-rank updates into
attention projections.  Both adapt a frozen model's behaviour but are
\emph{static} after training---they do not accumulate new information at
inference time.  Our memory adapter combines a trained read/write pathway
(like prefix tuning) with a persistent state that grows (like a KV cache),
yielding a system that is both trained and dynamic.

\paragraph{Attention-coupled latent memory.}
\citet{jeong2026lateral} introduced the $A^\top A V W$ write-back
operator for persistent lateralised memory banks and showed that
inhibitory cross-talk enables functional lateralization---an insight that
informs our understanding of why certain injection strategies succeed.
\citet{jeong2026brain} extended the framework with brain-region modules
(thalamic gating, hippocampal lateralization, amygdaloid salience,
prefrontal working memory), demonstrating that biological inductive
biases shape memory dynamics even in small-scale models.
\citet{jeong2026frozen} brought persistent memory to frozen
encoder--decoder LLMs, establishing the forgetting-curve evaluation
protocol that we adopt here.  The present paper extends that framework to
decoder-only models, completing the coverage of major architecture
families.

\section{Problem Setting}
\label{sec:setting}

We formalise the decoder-only persistent-memory problem in three steps:
first the stateless baseline that defines the control condition
(\S\ref{sec:baseline}), then the addition of a persistent memory bank
that converts a stateless model into a stateful one
(\S\ref{sec:adding_mem}), and finally the architectural challenge
specific to decoder-only models---the absence of a cross-attention
pathway (\S\ref{sec:decoder_challenge}).

\subsection{Stateless decoder-only baseline}
\label{sec:baseline}

A frozen decoder-only model processes the current turn
$x_t = (x_t^1, \ldots, x_t^n)$ autoregressively.  At each layer~$\ell$,
the hidden state $H^{(\ell)} \in \R^{n \times d}$ (sequence length~$n$,
model dimension~$d$) is transformed by causal self-attention:
\begin{align}\label{eq:selfattn}
  Q^{(\ell)} &= H^{(\ell)} W_Q^{(\ell)}, \quad
  K^{(\ell)} = H^{(\ell)} W_K^{(\ell)}, \quad
  V^{(\ell)} = H^{(\ell)} W_V^{(\ell)}, \notag\\
  A^{(\ell)} &= \softmax\!\biggl(\frac{Q^{(\ell)} {K^{(\ell)}}^\top}
    {\sqrt{d_k}} + M_{\mathrm{causal}}\biggr),
\end{align}
where $W_Q^{(\ell)}, W_K^{(\ell)} \in \R^{d \times d_k}$,
$W_V^{(\ell)} \in \R^{d \times d_v}$,
$Q^{(\ell)}, K^{(\ell)} \in \R^{n \times d_k}$,
$V^{(\ell)} \in \R^{n \times d_v}$,
$A^{(\ell)} \in \R^{n \times n}$,
and $M_{\mathrm{causal}} \in \R^{n \times n}$ is the causal mask.
All $W$ matrices are frozen.  The output $H_t \in \R^{n \times d}$
at the final layer is the model's only representation of the input;
it is discarded after generation.
Figure~\ref{fig:baseline} illustrates the stateless pipeline.

\begin{center}
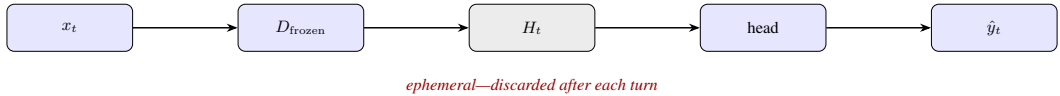

\begin{tikzpicture}[
  node distance=1.0cm and 2.0cm,
  block/.style={draw, rounded corners, minimum width=2.4cm,
                minimum height=0.9cm, align=center, font=\small},
  frozen/.style={block, fill=blue!10},
  latent/.style={block, fill=gray!15},
  arr/.style={-{Stealth[length=6pt]}, thick},
]
  \node[frozen] (x) {$x_t$};
  \node[frozen, right=of x] (dec) {$D_{\freeze}$};
  \node[latent, right=of dec] (H) {$H_t$};
  \node[frozen, right=of H] (head) {head};
  \node[frozen, right=of head] (y) {$\hat{y}_t$};

  \draw[arr] (x) -- (dec);
  \draw[arr] (dec) -- (H);
  \draw[arr] (H) -- (head);
  \draw[arr] (head) -- (y);

  \node[below=0.4cm of H, font=\small\itshape, text=red!60!black]
    {ephemeral---discarded after each turn};
\end{tikzpicture}

\captionof{figure}{Frozen decoder-only baseline.  The hidden state $H_t$ is
consumed within the current turn and then discarded; no information persists
across sessions.}
\label{fig:baseline}
\end{center}

\subsection{Adding persistent memory}
\label{sec:adding_mem}

We augment the stateless system with a persistent memory bank
$P_{t-1} \in \R^{n_P \times d}$ that survives across turns:
\begin{equation}\label{eq:stateful}
  H_t = D_{\freeze}(x_t;\, P_{t-1}), \qquad
  P_t = \mathrm{Write}(P_{t-1}, H_t), \qquad
  \hat{y}_t = \mathrm{head}(H_t).
\end{equation}
The semicolon in $D_{\freeze}(x_t;\, P_{t-1})$ indicates that $P_{t-1}$
is injected into the frozen forward pass through a method-specific read
path.  The $\mathrm{Write}$ operation updates memory from the current
hidden state.  The learned parameters~$\theta_{\mem}$ (read projections,
write projections, gates) are the only trainable weights; all original
model parameters remain frozen.

\subsection{The decoder-only challenge: no cross-attention}
\label{sec:decoder_challenge}

In encoder--decoder models, the decoder's cross-attention heads provide
a natural injection point: they were pre-trained to attend to
\emph{external} representations (encoder outputs), so re-purposing them
for memory entries requires minimal architectural disruption.

\begin{figure}[!hbtp]
\centering
\begin{tikzpicture}[
  node distance=0.6cm and 0.8cm,
  block/.style={draw, rounded corners, minimum width=1.8cm,
                minimum height=0.7cm, align=center, font=\scriptsize},
  frozen/.style={block, fill=blue!10},
  mem/.style={block, fill=green!15},
  gate/.style={block, fill=purple!15},
  arr/.style={-{Stealth[length=4pt]}, semithick},
  lbl/.style={font=\scriptsize\bfseries, text=black},
]
  \node[lbl] (labA) {\textbf{(a) KV Prefix}};
  \node[mem, below=0.3cm of labA] (PA) {$P_{t-1}$};
  \node[mem, right=0.6cm of PA] (kvA) {$K_\mem, V_\mem$};
  \node[frozen, right=0.8cm of kvA] (saA) {Self-Attn};
  \node[frozen, right=0.6cm of saA] (hA) {$H^{(\ell)\prime}$};
  \draw[arr] (PA) -- (kvA);
  \draw[arr] (kvA) -- (saA) node[midway, above, font=\tiny] {prepend};
  \draw[arr] (saA) -- (hA);

  \node[lbl, below=1.4cm of labA] (labB) {\textbf{(b) Parallel XAttn}};
  \node[frozen, below=0.3cm of labB] (saB) {Self-Attn};
  \node[mem, below=0.6cm of saB] (PB) {$P_{t-1}$};
  \node[mem, right=0.6cm of PB] (xaB) {XAttn$_\mem$};
  \node[frozen, right=0.6cm of saB] (sumB) {$+$};
  \node[frozen, right=0.6cm of sumB] (hB) {$H^{(\ell)\prime}$};
  \draw[arr] (saB) -- (sumB);
  \draw[arr] (PB) -- (xaB);
  \draw[arr] (xaB) -- (sumB) node[midway, right, font=\tiny] {$\beta$};
  \draw[arr] (sumB) -- (hB);

  \node[lbl, below=2.8cm of labB] (labC) {\textbf{(c) Gated Branch}};
  \node[frozen, below=0.3cm of labC] (saC) {Self-Attn};
  \node[mem, below=0.6cm of saC] (PC) {$P_{t-1}$};
  \node[mem, right=0.6cm of PC] (xaC) {XAttn$_\mem$};
  \node[gate, right=0.6cm of xaC] (gC) {$g_t \odot$};
  \node[frozen, right=0.6cm of saC] (sumC) {$+$};
  \node[frozen, right=0.6cm of sumC] (hC) {$H^{(\ell)\prime}$};
  \draw[arr] (saC) -- (sumC);
  \draw[arr] (PC) -- (xaC);
  \draw[arr] (xaC) -- (gC);
  \draw[arr] (gC) -- (sumC);
  \draw[arr] (sumC) -- (hC);
\end{tikzpicture}
\caption{Three decoder-only injection strategies.  (a)~KV prefix prepends
memory-derived keys and values to the self-attention cache.
(b)~Parallel cross-attention inserts a new attention pathway alongside
the existing self-attention.  (c)~Gated branch computes a memory readout
and adds it through a learned content-dependent gate.}
\label{fig:three_strategies}
\end{figure}
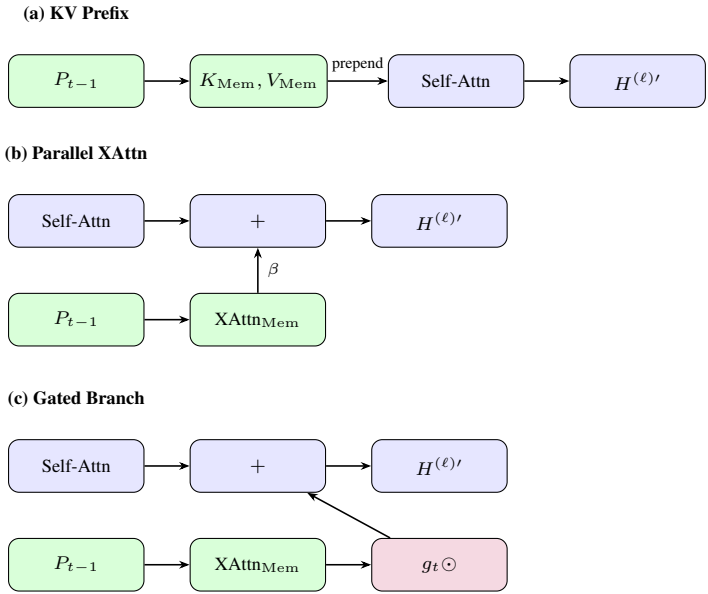

Decoder-only models have no such pathway.  Their self-attention heads
were pre-trained to attend exclusively to preceding tokens within the
same sequence.  Injecting external persistent state requires one of three
strategies (illustrated in Figure~\ref{fig:three_strategies}):

\begin{enumerate}[leftmargin=*,itemsep=2pt]
\item \textbf{KV prefix.}
  Project $P$ into additional key--value pairs and prepend them to each
  layer's self-attention.  The model ``sees'' memory as if it were earlier
  tokens.  This is the most natural analog of how decoder-only models
  already process context.

\item \textbf{Parallel cross-attention insertion.}
  Insert new cross-attention layers alongside the existing self-attention
  (Flamingo-style).  More expressive but heavier: each inserted layer
  adds a full set of $Q$, $K$, $V$, $O$ projections.

\item \textbf{Gated additive branch.}
  Compute a memory readout and add it to the hidden state through a
  learned gate, without modifying the self-attention computation itself.
\end{enumerate}

Table~\ref{tab:injection} maps the six methods to their decoder-only
read strategies.

\begin{table}[!hbtp]
\centering
\caption{Mapping of the six memory methods from the encoder--decoder
setting~\citep{jeong2026frozen} to the decoder-only setting.  The write
rule is identical; only the read injection point changes.}
\label{tab:injection}
\small
\begin{tabular}{@{}llll@{}}
\toprule
Method & Enc-Dec read path & Decoder-only read path & Strategy \\
\midrule
M.1 Prefix      & Encoder-input prefix   & Self-attention KV prefix       & KV prefix \\
M.2 XAttn       & Parallel decoder XAttn & Parallel cross-attention       & Inserted XAttn \\
M.3 KV Ext      & Decoder KV extension   & Self-attention KV extension    & KV prefix \\
M.4 Hebbian     & Decoder KV extension   & Self-attention KV extension    & KV prefix \\
M.5 Gated       & Context-gated decoder  & Gated additive branch          & Gated branch \\
M.6 Slot        & Decoder KV extension   & Self-attention KV extension    & KV prefix \\
\bottomrule
\end{tabular}
\end{table}

\section{Adapting the Six Methods to Decoder-Only Models}
\label{sec:methods}

The write rule for all methods is architecture-agnostic: it takes the
current hidden state $H_t$ and the previous memory $P_{t-1}$ and
produces $P_t$.  We reproduce each write rule verbatim
from~\citet{jeong2026frozen} and describe only the decoder-only read
adaptations below.

\subsection{Shared write rule}

For methods M.1--M.3, M.5, and M.6, the write rule is the
attention-coupled update:
\begin{align}\label{eq:write}
  Q_w &= H_t W_Q^w, \quad
  K_w = P_{t-1} W_K^w, \quad
  V_w = H_t W_V^w, \notag\\
  A_t &= \softmax\!\biggl(\frac{Q_w K_w^\top}{\sqrt{d}}\biggr), \quad
  P_t = \gamma\, P_{t-1} + A_t^\top V_w.
\end{align}
Here $W_Q^w, W_V^w \in \R^{d \times d}$ project the hidden state and
$W_K^w \in \R^{d \times d}$ projects memory, yielding
$Q_w, V_w \in \R^{n \times d}$, $K_w \in \R^{n_P \times d}$,
and $A_t \in \R^{n \times n_P}$.
Values $V_w$ are drawn from $H_t$ (not from $P$), ensuring new content
enters memory regardless of $P$'s current state.  For M.4 (Hebbian),
the write is an outer-product rule; for M.6 (Slot), a sparse top-$k$
overwrite.  These are unchanged from the encoder--decoder setting.

\subsection{M.1: Self-Attention KV Prefix}
\label{sec:m1}

Memory $P_{t-1} \in \R^{n_P \times d}$ is projected into soft key--value pairs and prepended
to every self-attention layer:
\begin{align}\label{eq:m1_read}
  K_{\mem}^{(\ell)} &= P_{t-1}\, W_{K,\mem}^{(\ell)}, \quad
  V_{\mem}^{(\ell)} = P_{t-1}\, W_{V,\mem}^{(\ell)}, \notag\\
  \tilde{K}^{(\ell)} &= \bigl[K_{\mem}^{(\ell)};\; K^{(\ell)}\bigr], \quad
  \tilde{V}^{(\ell)} = \bigl[V_{\mem}^{(\ell)};\; V^{(\ell)}\bigr].
\end{align}
Here $W_{K,\mem}^{(\ell)} \in \R^{d \times d_k}$ and
$W_{V,\mem}^{(\ell)} \in \R^{d \times d_v}$ are learnable projections,
yielding $K_{\mem}^{(\ell)} \in \R^{n_P \times d_k}$,
$V_{\mem}^{(\ell)} \in \R^{n_P \times d_v}$,
and concatenated tensors
$\tilde{K}^{(\ell)} \in \R^{(n_P+n) \times d_k}$,
$\tilde{V}^{(\ell)} \in \R^{(n_P+n) \times d_v}$.
The causal mask is extended: memory positions are visible to all input
tokens (no causal restriction), while the original causal structure among
input tokens is preserved.  The query $Q^{(\ell)}$ is unchanged.  This
is the most natural decoder-only injection: the model treats memory
entries as if they were earlier tokens in the sequence.
Figure~\ref{fig:m1} illustrates the KV prefix injection.

\begin{center}
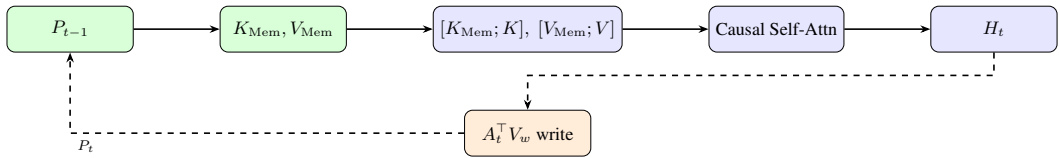

\begin{tikzpicture}[
  node distance=0.9cm and 1.5cm,
  block/.style={draw, rounded corners, minimum width=2.2cm,
                minimum height=0.8cm, align=center, font=\small},
  frozen/.style={block, fill=blue!10},
  mem/.style={block, fill=green!15},
  write/.style={block, fill=orange!15},
  arr/.style={-{Stealth[length=5pt]}, thick},
  darr/.style={-{Stealth[length=5pt]}, thick, dashed},
]
  \node[mem] (P) {$P_{t-1}$};
  \node[mem, right=of P] (proj) {$K_\mem, V_\mem$};
  \node[frozen, right=of proj] (cat) {$[K_\mem; K],\;[V_\mem; V]$};
  \node[frozen, right=of cat] (sa) {Causal Self-Attn};
  \node[frozen, right=of sa] (H) {$H_t$};
  \node[write, below=1.0cm of cat] (write) {$A_t^\top V_w$ write};

  \draw[arr] (P) -- (proj);
  \draw[arr] (proj) -- (cat);
  \draw[arr] (cat) -- (sa);
  \draw[arr] (sa) -- (H);
  \draw[darr] (H.south) -- ++(0,-0.4) -| (write);
  \draw[darr] (write) -| node[below right, font=\scriptsize]{$P_t$} (P);
\end{tikzpicture}

\captionof{figure}{M.1: Memory is projected into soft key--value pairs and
prepended to the self-attention cache at every layer.  The write rule
updates $P$ from $H_t$.}
\label{fig:m1}
\end{center}

\subsection{M.2: Inserted Parallel Cross-Attention}
\label{sec:m2}

A new cross-attention layer is inserted after each frozen self-attention
block, attending from the hidden state to the memory bank:
\begin{equation}\label{eq:m2_read}
  c_{\mem}^{(\ell)} = \mathrm{XAttn}_{\mem}^{(\ell)}\!\bigl(H^{(\ell)},\, P_{t-1}\bigr),
  \qquad
  {H^{(\ell)}}' = H^{(\ell)} + \beta^{(\ell)}\, c_{\mem}^{(\ell)},
\end{equation}
where $c_{\mem}^{(\ell)} \in \R^{n \times d}$ is the cross-attention
output and $\beta^{(\ell)} \in \R$ is a learnable scalar initialised to
zero, so the model starts as the unmodified frozen baseline.  This follows the
Flamingo design~\citep{alayrac2022flamingo} and is the only method that
introduces a genuinely new attention pathway into the decoder-only model.
Figure~\ref{fig:m2} shows the parallel insertion.

\begin{center}
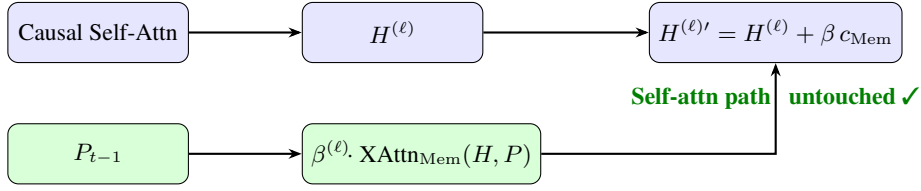

\begin{tikzpicture}[
  node distance=0.9cm and 1.5cm,
  block/.style={draw, rounded corners, minimum width=2.4cm,
                minimum height=0.8cm, align=center, font=\small},
  frozen/.style={block, fill=blue!10},
  mem/.style={block, fill=green!15},
  arr/.style={-{Stealth[length=5pt]}, thick},
]
  \node[frozen] (sa) {Causal Self-Attn};
  \node[frozen, right=of sa] (Hl) {$H^{(\ell)}$};
  \node[mem, below=0.8cm of sa] (P) {$P_{t-1}$};
  \node[mem, right=of P] (memx)
    {$\beta^{(\ell)}\!\cdot$ XAttn$_{\mem}(H, P)$};
  \node[frozen, right=2.2cm of Hl] (sum) {$H^{(\ell)\prime} = H^{(\ell)} + \beta\, c_{\mem}$};

  \draw[arr] (sa) -- (Hl);
  \draw[arr] (Hl) -- (sum);
  \draw[arr] (P) -- (memx);
  \draw[arr] (memx) -| (sum);

  \node[below=0.2cm of sum, text=green!50!black, font=\small\bfseries]
    {Self-attn path ~~untouched \ding{51}};
\end{tikzpicture}

\captionof{figure}{M.2: A parallel cross-attention layer is inserted after
each frozen self-attention block.  The scaling factor $\beta^{(\ell)}$ is
initialised to zero for safe startup.}
\label{fig:m2}
\end{center}

\subsection{M.3: Self-Attention KV Extension}
\label{sec:m3}

M.3 uses the same KV extension read path as M.1---projecting $P_{t-1}$
into key--value pairs and concatenating them with the frozen
self-attention---but applies the attention-coupled write rule
(Eq.~\ref{eq:write}) with \emph{per-layer} projections.  Each frozen
layer~$\ell$ maintains its own read projection pair:
\begin{align}\label{eq:m3_write}
  K_{\mathrm{ext}}^{(\ell)} &= P_{t-1}\, W_{K,\mathrm{ext}}^{(\ell)}, \quad
  V_{\mathrm{ext}}^{(\ell)} = P_{t-1}\, W_{V,\mathrm{ext}}^{(\ell)},
\end{align}
where $W_{K,\mathrm{ext}}^{(\ell)} \in \R^{d \times d_k}$ and
$W_{V,\mathrm{ext}}^{(\ell)} \in \R^{d \times d_v}$ are trainable, yielding
$K_{\mathrm{ext}}^{(\ell)} \in \R^{n_P \times d_k}$ and
$V_{\mathrm{ext}}^{(\ell)} \in \R^{n_P \times d_v}$.
These are concatenated with the frozen self-attention keys and values:
\begin{align}\label{eq:m3_read}
  \tilde{K}^{(\ell)} &= \bigl[K_{\mathrm{ext}}^{(\ell)};\; K^{(\ell)}\bigr], \quad
  \tilde{V}^{(\ell)} = \bigl[V_{\mathrm{ext}}^{(\ell)};\; V^{(\ell)}\bigr],
\end{align}
producing $\tilde{K}^{(\ell)} \in \R^{(n_P+n) \times d_k}$ and
$\tilde{V}^{(\ell)} \in \R^{(n_P+n) \times d_v}$.
The causal mask extension and query $Q^{(\ell)}$ are identical to M.1
(Eq.~\ref{eq:m1_read}).

The key difference from M.1 is that M.3's write update is also applied
per-layer: the attention-coupled rule
$P_t^{(\ell)} = \gamma P_{t-1}^{(\ell)} + {A_t^{(\ell)}}^\top V_w^{(\ell)}$
uses layer-specific projections, so each transformer layer can write to
and read from its own view of the memory bank.  M.1, by contrast, shares
a single global write across all layers.
Figure~\ref{fig:m3} illustrates the per-layer KV extension.

\begin{center}
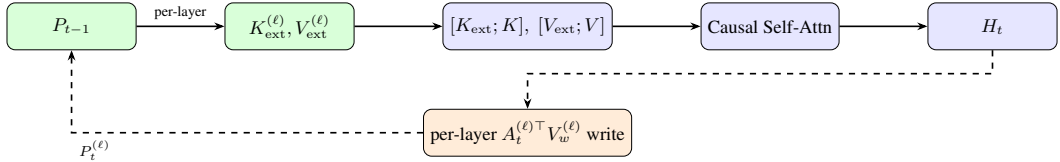

\begin{tikzpicture}[
  node distance=0.9cm and 1.5cm,
  block/.style={draw, rounded corners, minimum width=2.2cm,
                minimum height=0.8cm, align=center, font=\small},
  frozen/.style={block, fill=blue!10},
  mem/.style={block, fill=green!15},
  write/.style={block, fill=orange!15},
  arr/.style={-{Stealth[length=5pt]}, thick},
  darr/.style={-{Stealth[length=5pt]}, thick, dashed},
]
  \node[mem] (P) {$P_{t-1}$};
  \node[mem, right=of P] (proj) {$K_{\mathrm{ext}}^{(\ell)}, V_{\mathrm{ext}}^{(\ell)}$};
  \node[frozen, right=of proj] (cat) {$[K_{\mathrm{ext}}; K],\;[V_{\mathrm{ext}}; V]$};
  \node[frozen, right=of cat] (sa) {Causal Self-Attn};
  \node[frozen, right=of sa] (H) {$H_t$};
  \node[write, below=1.0cm of cat] (write) {per-layer $A_t^{(\ell)\top} V_w^{(\ell)}$ write};

  \draw[arr] (P) -- (proj) node[midway, above, font=\scriptsize] {per-layer};
  \draw[arr] (proj) -- (cat);
  \draw[arr] (cat) -- (sa);
  \draw[arr] (sa) -- (H);
  \draw[darr] (H.south) -- ++(0,-0.4) -| (write);
  \draw[darr] (write) -| node[below right, font=\scriptsize]{$P_t^{(\ell)}$} (P);
\end{tikzpicture}

\captionof{figure}{M.3: Memory is projected into per-layer key--value pairs
and concatenated with the frozen self-attention cache.  Unlike M.1, both
the read projections and the write update are layer-specific.}
\label{fig:m3}
\end{center}

\subsection{M.4: Hebbian / Associative Memory}
\label{sec:m4}

The Hebbian matrix $M_t \in \R^{d_h \times d_h}$ is queried by the
current hidden state to produce a recall vector, which is then exposed
to the decoder as additional KV entries:
\begin{equation}\label{eq:m4_read}
  R_t = (H_t\, W_{Q,H})\, M_{t-1}, \quad
  K_M^{(\ell)} = R_t\, W_{K,\mem}^{(\ell)}, \quad
  V_M^{(\ell)} = R_t\, W_{V,\mem}^{(\ell)},
\end{equation}
where $W_{Q,H} \in \R^{d \times d_h}$ projects the hidden state into
the associative space, yielding $R_t \in \R^{n \times d_h}$.
The per-layer projections
$W_{K,\mem}^{(\ell)} \in \R^{d_h \times d_k}$ and
$W_{V,\mem}^{(\ell)} \in \R^{d_h \times d_v}$ produce
$K_M^{(\ell)} \in \R^{n \times d_k}$ and
$V_M^{(\ell)} \in \R^{n \times d_v}$,
concatenated to the self-attention keys and values as in M.1/M.3.
Figure~\ref{fig:m4} illustrates the Hebbian read path.

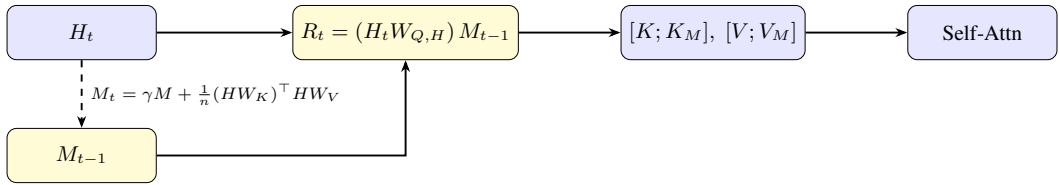
\begin{center}
\begin{tikzpicture}[
  node distance=0.9cm and 1.5cm,
  block/.style={draw, rounded corners, minimum width=2.2cm,
                minimum height=0.8cm, align=center, font=\small},
  frozen/.style={block, fill=blue!10},
  mem/.style={block, fill=yellow!20},
  arr/.style={-{Stealth[length=5pt]}, thick},
  darr/.style={-{Stealth[length=5pt]}, thick, dashed},
]
  \node[frozen] (H) {$H_t$};
  \node[mem, below=1.0cm of H] (M) {$M_{t-1}$};
  \node[mem, right=2.0cm of H] (read) {$R_t = (H_t W_{Q,H})\, M_{t-1}$};
  \node[frozen, right=of read] (kv) {$[K;K_M],\; [V;V_M]$};
  \node[frozen, right=of kv] (sa) {Self-Attn};

  \draw[arr] (H) -- (read);
  \draw[arr] (M) -| (read);
  \draw[arr] (read) -- (kv);
  \draw[arr] (kv) -- (sa);
  \draw[darr] (H) -- (M) node[midway, right, font=\scriptsize]
    {$M_t = \gamma M + \frac{1}{n}(HW_K)^\top HW_V$};
\end{tikzpicture}

\captionof{figure}{M.4: The Hebbian matrix is queried by the current hidden
state; the recalled vector is projected into additional KV pairs for
self-attention.}
\label{fig:m4}
\end{center}

\subsection{M.5: Gated Additive Branch}
\label{sec:m5}

A lightweight memory branch reads from $P$ and contributes to the hidden
state through a content-dependent gate:
\begin{align}\label{eq:m5_read}
  c_{\mem}^{(\ell)} &= \mathrm{XAttn}_{\mem}^{(\ell)}\!\bigl(H^{(\ell)},\, P_{t-1}\bigr), \notag\\
  g_t^{(\ell)} &= \sigma\!\bigl(W_g^{(\ell)}[H^{(\ell)};\, c_{\mem}^{(\ell)}] + b_g^{(\ell)}\bigr), \notag\\
  {H^{(\ell)}}' &= H^{(\ell)} + g_t^{(\ell)} \odot c_{\mem}^{(\ell)}.
\end{align}
Here $c_{\mem}^{(\ell)} \in \R^{n \times d}$ is the memory readout,
$W_g^{(\ell)} \in \R^{d \times 2d}$ maps the concatenation
$[H^{(\ell)};\, c_{\mem}^{(\ell)}] \in \R^{n \times 2d}$ through a sigmoid
to produce the gate $g_t^{(\ell)} \in \R^{n \times d}$, and
$b_g^{(\ell)} \in \R^{d}$ is the gate bias.
With $b_g^{(\ell)} < 0$ at initialisation, the gate starts nearly closed
(safe startup).  Figure~\ref{fig:m5} illustrates the gated branch.

\begin{center}
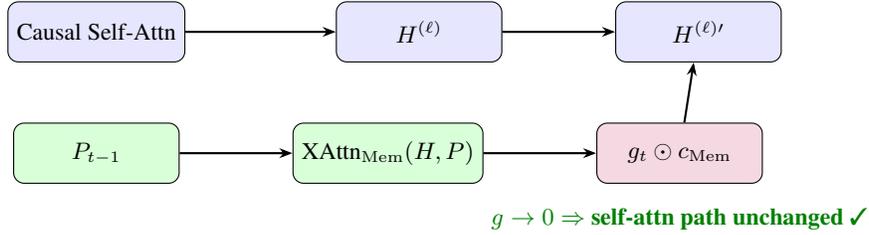

\begin{tikzpicture}[
  node distance=0.9cm and 1.5cm,
  block/.style={draw, rounded corners, minimum width=2.2cm,
                minimum height=0.8cm, align=center, font=\small},
  frozen/.style={block, fill=blue!10},
  mem/.style={block, fill=green!15},
  gate/.style={block, fill=purple!15},
  arr/.style={-{Stealth[length=5pt]}, thick},
]
  \node[frozen] (sa) {Causal Self-Attn};
  \node[frozen, right=2.0cm of sa] (base) {$H^{(\ell)}$};
  \node[mem, below=0.8cm of sa] (P) {$P_{t-1}$};
  \node[mem, right=of P] (read) {XAttn$_{\mem}(H,P)$};
  \node[gate, right=of read] (gate) {$g_t \odot c_{\mem}$};
  \node[frozen, right=of base] (sum) {$H^{(\ell)\prime}$};

  \draw[arr] (sa) -- (base);
  \draw[arr] (base) -- (sum);
  \draw[arr] (P) -- (read);
  \draw[arr] (read) -- (gate);
  \draw[arr] (gate) -- (sum);

  \node[below=0.2cm of gate, text=green!50!black, font=\small\bfseries]
    {$g \to 0 \Rightarrow$ self-attn path unchanged \ding{51}};
\end{tikzpicture}

\captionof{figure}{M.5: A gated memory branch reads from $P$ and contributes
through a content-dependent gate.  At initialisation $b_g < 0$ keeps the
gate nearly closed, preserving the frozen self-attention path.}
\label{fig:m5}
\end{center}

\subsection{M.6: Slot-Based Sparse Write with KV Read}
\label{sec:m6}

The memory bank $P \in \R^{S \times d}$ is organised as $S$ addressable
slots.  The write rule uses sparse top-$k$ addressing to selectively
overwrite a small subset of slots each turn, preventing the dilution
that affects dense write rules when capacity is limited.

\paragraph{Addressing.}
A learnable address head computes a similarity score between each
hidden-state token and every slot:
\begin{equation}\label{eq:m6_address}
  a_{ij} = \frac{(H_t\, W_A)_i \cdot (P_{t-1}\, W_S)_j}{\sqrt{d}},
  \qquad
  \alpha_j = \max_i\, a_{ij},
\end{equation}
where $W_A, W_S \in \R^{d \times d}$ are learnable projections,
$a_{ij}$ is the affinity between token~$i$ and slot~$j$, and
$\alpha_j \in \R$ is the maximum affinity over all tokens for slot~$j$.
The top-$k$ slots by $\alpha_j$ are selected for writing.

\paragraph{Sparse write.}
Let $\mathcal{T} = \topk_j(\alpha_j)$ be the index set of the $k$
highest-scored slots.  Each selected slot is updated with a gated blend
of its previous content and the new value aggregated from the hidden
state:
\begin{equation}\label{eq:m6_write}
  P_t^{(j)} =
  \begin{cases}
    \gamma\, P_{t-1}^{(j)} + (1 - \gamma)\, v_j & \text{if } j \in \mathcal{T}, \\
    P_{t-1}^{(j)} & \text{otherwise},
  \end{cases}
\end{equation}
where $v_j = \sum_i \softmax(a_{ij})\,(H_t W_V^w)_i \in \R^d$ is the
attention-weighted aggregation of hidden-state values into slot~$j$, and
$\gamma$ is the shared decay factor.  Only $k$ of $S$ slots change each turn,
giving the write rule a discrete selection mechanism that prevents dilution.

\paragraph{KV read.}
The read path exposes all $S$ slots to the decoder via self-attention KV
extension, identical to M.1/M.3:
\begin{equation}\label{eq:m6_read}
  K_{\mem}^{(\ell)} = P_t\, W_{K,\mem}^{(\ell)}, \quad
  V_{\mem}^{(\ell)} = P_t\, W_{V,\mem}^{(\ell)}, \quad
  \tilde{K}^{(\ell)} = [K_{\mem}^{(\ell)};\, K^{(\ell)}], \quad
  \tilde{V}^{(\ell)} = [V_{\mem}^{(\ell)};\, V^{(\ell)}].
\end{equation}
Figure~\ref{fig:m6} illustrates the slot-based design.

\begin{center}
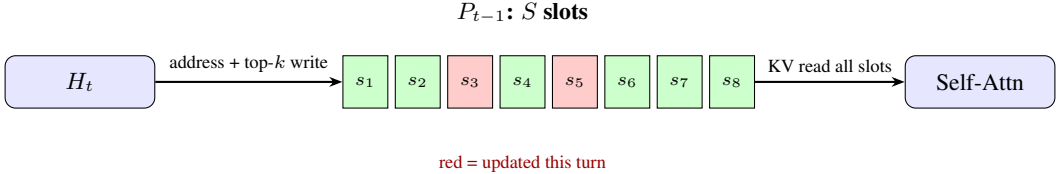

\begin{tikzpicture}[
  node distance=0.8cm and 1.4cm,
  block/.style={draw, rounded corners, minimum width=2.0cm,
                minimum height=0.7cm, align=center, font=\small},
  frozen/.style={block, fill=blue!10},
  slot/.style={draw, minimum width=0.6cm, minimum height=0.7cm,
               font=\scriptsize, align=center},
  arr/.style={-{Stealth[length=5pt]}, thick},
]
  \node[frozen] (H) {$H_t$};
  \node[right=2.5cm of H] (slots) {};

  \foreach \i/\c in {1/green!20, 2/green!20, 3/red!20, 4/green!20, 5/red!20,
                     6/green!20, 7/green!20, 8/green!20} {
    \node[slot, fill=\c, right=\i*0.7cm-0.7cm of slots.west] (s\i) {$s_\i$};
  }
  \node[above=0.3cm of s4, font=\small\bfseries] {$P_{t-1}$: $S$ slots};
  \node[frozen, right=2.0cm of s8] (sa) {Self-Attn};

  \draw[arr] (H) -- (s1) node[midway, above, font=\scriptsize] {address + top-$k$ write};
  \draw[arr] (s8) -- (sa) node[midway, above, font=\scriptsize] {KV read all slots};

  \node[below=0.5cm of s4, font=\scriptsize, text=red!60!black]
    {red = updated this turn};
\end{tikzpicture}

\captionof{figure}{M.6: Memory is organised as $S$ addressable slots.
Only the top-$k$ addressed slots are updated each turn (sparse write);
all slots are read via self-attention KV extension.}
\label{fig:m6}
\end{center}

\section{Training and Inference}
\label{sec:training}

The training protocol mirrors~\citet{jeong2026frozen} exactly.

\paragraph{Type~1: Supervised learning.}
All adapter parameters $\theta_{\mem}$ are trained while the decoder
backbone remains frozen.  Gradients flow through the frozen model (which
acts as a fixed differentiable function) to reach $\theta_{\mem}$:
\begin{equation}\label{eq:train}
  \theta_{\mem} \leftarrow
    \theta_{\mem} - \eta\,\nabla_{\theta_{\mem}}
    \mathcal{L}\!\bigl(D_{\freeze}(x_t;\, P_{t-1}),\; y_t\bigr).
\end{equation}
Write-side projections execute without gradients (detached) to prevent
the computation graph from growing across the full conversation history.
They act as fixed random maps that preserve pairwise distances
(Johnson--Lindenstrauss property).

\paragraph{Type~2: Conversational learning.}
After training, $\theta_{\mem}$ is frozen but $P_t$ continues to
accumulate at inference time without gradients:
$P_t = \mathrm{Write}(P_{t-1}, H_t)$ with $\theta_{\mem}$ fixed.
Each new conversation enriches~$P$, and the system becomes more informed
with every turn---driven by ordinary dialogue rather than curated
datasets.

\paragraph{Implementation details.}
We use GPT-2 (124M parameters) as the frozen backbone in bfloat16.  Memory adapters are trained with AdamW
(learning rate $10^{-4}$, weight decay $10^{-2}$, linear warmup of 200
steps, gradient norm clipped at 1.0) for 10 epochs with batch size~4
and gradient accumulation~4 (effective batch~16).
Shared memory hyperparameters: bank size $n_P = 64$, write
decay $\gamma = 0.95$, $d_h = 256$ (M.4), $S = 64$ slots with
top-$k = 8$ writes (M.6)---all identical to the encoder--decoder setting.
Truncated backpropagation uses a window of $k = 8$ turns.
Early stopping halts training if validation loss does not improve for
3~consecutive epochs.
Figure~\ref{fig:learning_loops} contrasts the two phases.

\begin{center}
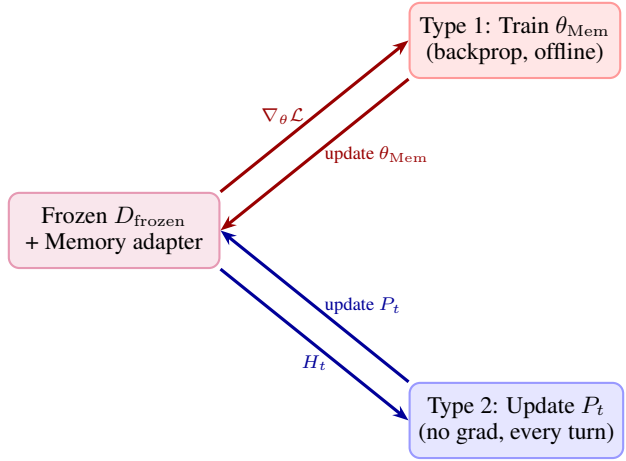

\begin{tikzpicture}[
  node distance=1.2cm and 2.0cm,
  block/.style={draw, rounded corners, minimum width=2.8cm,
                minimum height=1.0cm, align=center, font=\small},
  type1/.style={block, fill=red!10, thick, draw=red!40},
  type2/.style={block, fill=blue!10, thick, draw=blue!40},
  both/.style={block, fill=purple!10, thick, draw=purple!40},
  arr/.style={-{Stealth[length=6pt]}, very thick},
  t1arr/.style={-{Stealth[length=6pt]}, very thick, red!60!black},
  t2arr/.style={-{Stealth[length=6pt]}, very thick, blue!60!black},
]
  \node[both] (model) {Frozen $D_{\freeze}$\\+ Memory adapter};
  \node[type1, above right=1.5cm and 2.5cm of model] (t1)
    {Type~1: Train $\theta_{\mem}$\\(backprop, offline)};
  \node[type2, below right=1.5cm and 2.5cm of model] (t2)
    {Type~2: Update $P_t$\\(no grad, every turn)};

  \draw[t1arr] (model.north east) -- (t1.west)
    node[midway, left, font=\scriptsize, text=red!60!black]
    {$\nabla_\theta \mathcal{L}$};
  \draw[t1arr] (t1.south west) -- (model.east)
    node[midway, right, font=\scriptsize, text=red!60!black]
    {update $\theta_{\mem}$};
  \draw[t2arr] (model.south east) -- (t2.west)
    node[midway, below, font=\scriptsize, text=blue!60!black]
    {$H_t$};
  \draw[t2arr] (t2.north west) -- (model.east)
    node[midway, right, font=\scriptsize, text=blue!60!black]
    {update $P_t$};
\end{tikzpicture}

\captionof{figure}{Two learning phases.  Type~1 trains the memory adapter
parameters offline by backpropagation through the frozen decoder.
Type~2 updates the persistent memory online at each conversation turn
with all model weights frozen.}
\label{fig:learning_loops}
\end{center}

Table~\ref{tab:master} consolidates the six methods along key design
dimensions for the decoder-only setting.

\begin{table}[!hbtp]
\centering
\caption{Six trained persistent-memory methods compared for decoder-only
models.  ``Self-attn safe'' means the original frozen self-attention route
is preserved.  ``Memory cost'' is per turn.}
\label{tab:master}
\renewcommand{\arraystretch}{1.15}
\resizebox{\textwidth}{!}{%
\begin{tabular}{clccccl}
\toprule
& \textbf{Method}
  & \textbf{Injection strategy}
  & \textbf{Self-attn safe}
  & \textbf{New params}
  & \textbf{Mem.\ cost}
  & \textbf{Write mechanism} \\
\midrule
1 & Self-attention KV prefix
  & KV prefix & \ding{51} & $\sim$2.4M & const.
  & $A_t^\top V$ \\
2 & Parallel cross-attention
  & Inserted XAttn & \ding{51} & $\sim$4.1M & const.
  & $A_t^\top V$ \\
3 & Self-attention KV extension
  & KV prefix & \ding{51} & $\sim$2.4M & const.
  & $A_t^\top V$ (per-layer) \\
4 & Hebbian / associative
  & KV prefix & \ding{51} & $\sim$0.8M & $O(d_h^2)$
  & Hebbian outer prod.\ \\
5 & Gated additive branch
  & Gated branch & \ding{51} & $\sim$4.7M & const.
  & gated XAttn \\
6 & Slot-based sparse write
  & KV prefix & \ding{51} & $\sim$1.8M & $O(Sd)$
  & Top-$k$ overwrite \\
\bottomrule
\end{tabular}}%
\end{table}

\section{Evaluation}
\label{sec:eval}

The preceding sections specify \emph{how} each method reads and writes
persistent memory; this section specifies \emph{what} we measure and
\emph{how} we measure it.  We adopt the absolute forgetting-curve
protocol from~\citet{jeong2026frozen} without modification, apply the
same equal-input principle, and extend the evaluation to a
cross-architecture comparison spanning encoder--decoder and decoder-only
backbones.

\subsection{Forgetting-curve protocol}

We adopt the absolute forgetting-curve evaluation
from~\citet{jeong2026frozen} without modification.  For a question~$q$
asked after $T$ conversational turns with annotated evidence turns
$\mathcal{E}_q$, the evidence lag is:
\begin{equation}\label{eq:lag}
  \ell_q = T - \min(\mathcal{E}_q).
\end{equation}
The retained-memory score measures how much answer quality disappears
when a method's own persistent state is ablated:
\begin{equation}\label{eq:retained}
  R_q(m) = \max\!\bigl(0,\;
    F_1(\hat{y}_q^{\mem}(m),\, y_q) - F_1(\hat{y}_q^{0}(m),\, y_q)\bigr).
\end{equation}
The stateless baseline has $R_q = 0$ by construction.  For
decoder-only evaluation we use five lag buckets:
$[0, 32)$, $[32, 64)$, $[64, 128)$, $[128, 256)$, and $[256, \infty)$
turns, averaged within each bucket, and smoothed by a weighted
non-increasing isotonic fit (PAVA), following the five-bucket scheme
from~\citet{jeong2026frozen}.

\subsection{Equal-input principle}

Every condition---baseline and all six memory methods on each
backbone---receives exactly the same input $x_t$ at each turn: the
current conversational turn only.  No method receives the full history.
Any non-zero retained-memory score is attributable solely to the
persistent state~$P$.

\subsection{Benchmark}

The primary benchmark is LoCoMo~\citep{maharana2024locomo}, a long-term
conversational-memory dataset with explicit QA supervision and annotated
evidence turns.

\subsection{Knowledge-accumulation protocol}

We complement the forgetting curve with a cumulative knowledge
evaluation over LoCoMo's 30~sessions.  At each session boundary~$s$, we
measure the knowledge score~$K_s$: the F1 improvement on factual
questions about sessions $0, \ldots, s$ relative to the stateless
baseline.  The overall metric $\Delta K = K_{29}$ captures total
accumulated knowledge at the final session.  Positive $\Delta K$ means
the model has learned facts from the conversation that the stateless
baseline cannot access.

\subsection{Capacity conditions}
\label{sec:capacity_conditions}

A central question is whether the inductive bias of a memory method or
its raw capacity determines success.  To disentangle the two, we
evaluate every method under two capacity budgets that share all other
hyperparameters (backbone, training schedule, decay~$\gamma$).

The \textbf{standard ($1\times$) setting} uses the same hyperparameters
as the encoder--decoder experiments of~\citet{jeong2026frozen}: memory
bank size $n_P = 64$, Hebbian dimension $d_h = 256$, slot count $S = 64$,
and top-$k = 8$ writes.  This is a deliberately constrained regime:
with 64 memory vectors the adapter must compress an entire conversation
history into roughly 49\,K parameters of persistent state.  Any method
that succeeds here must rely on a strong inductive bias to route the
right information into the limited capacity.

The \textbf{large ($10\times$) setting} scales every capacity parameter
by a factor of ten: $n_P = 640$, $d_h = 810$, $S = 640$, top-$k = 80$.
If a method fails at $1\times$ but succeeds at $10\times$, the failure
is attributable to insufficient capacity rather than a fundamental
incompatibility with the decoder-only injection path.  Conversely, if a
method already saturates at $1\times$, extra capacity yields diminishing
returns---confirming that its inductive bias is already sufficient.

Table~\ref{tab:capacity} summarises the two conditions.

\begin{table}[!hbtp]
\centering
\caption{Memory capacity conditions.  All other hyperparameters
(backbone, learning rate, decay~$\gamma$, training epochs) are shared.}
\label{tab:capacity}
\small
\begin{tabular}{@{}lcc@{}}
\toprule
\textbf{Parameter} & $\mathbf{1\times}$ & $\mathbf{10\times}$ \\
\midrule
Memory bank size $n_P$   & 64  & 640 \\
Hebbian dimension $d_h$  & 256 & 810 \\
Slot count $S$           & 64  & 640 \\
Top-$k$ writes           & 8   & 80  \\
\bottomrule
\end{tabular}
\end{table}

\section{Results}
\label{sec:results}

We organise the empirical findings around four questions: which methods
produce positive forgetting curves at $1\times$ capacity
(\S\ref{sec:curves})?  What separates the methods that succeed from
those that fail (\S\ref{sec:dichotomy})?  Does $10\times$ capacity
bridge the gap (\S\ref{sec:capacity})?  And does knowledge accumulate
across sessions (\S\ref{sec:knowledge})?

\subsection{Forgetting curves at $1\times$ capacity}
\label{sec:curves}

\begin{figure}[!hbtp]
\centering
\begin{tikzpicture}
\begin{axis}[
    width=0.92\linewidth,
    height=6cm,
    xlabel={Evidence lag (turns in the past)},
    ylabel={Retained-memory score (\%)},
    xmin=0.5, xmax=5.5,
    ymin=0, ymax=20,
    xtick={1,2,3,4,5},
    xticklabels={{$0$--$31$},{$32$--$63$},{$64$--$127$},{$128$--$255$},{$256+$}},
    x tick label style={rotate=25, anchor=east, font=\scriptsize},
    legend style={
      at={(1.02,0.5)},
      anchor=west,
      font=\scriptsize,
      cells={anchor=west},
    },
    grid=major,
    grid style={gray!30},
    every axis plot/.append style={thick},
    tick label style={font=\small},
    label style={font=\small},
    title={\textbf{GPT-2 ($1\times$ capacity)}},
]
\addplot[black, dashed] coordinates {(1,0) (2,0) (3,0) (4,0) (5,0)};
\addlegendentry{Baseline}
\addplot[red, mark=triangle*] coordinates {(1,17.85) (2,14.65) (3,9.02) (4,9.02) (5,9.02)};
\addlegendentry{M.2 XAttn}
\addplot[olive, mark=pentagon*] coordinates {(1,9.51) (2,9.51) (3,9.23) (4,9.23) (5,9.23)};
\addlegendentry{M.4 Hebbian}
\addplot[orange, mark=otimes*] coordinates {(1,17.21) (2,13.91) (3,7.08) (4,7.08) (5,7.08)};
\addlegendentry{M.6 Slot}
\addplot[purple, mark=star] coordinates {(1,0.36) (2,0.10) (3,0.10) (4,0.10) (5,0.09)};
\addlegendentry{M.5 Gated}
\addplot[blue, mark=square*] coordinates {(1,0.02) (2,0.02) (3,0.02) (4,0.02) (5,0.02)};
\addlegendentry{M.1 Prefix}
\addplot[teal, mark=diamond*] coordinates {(1,0.0) (2,0.0) (3,0.0) (4,0.0) (5,0.0)};
\addlegendentry{M.3 KV Ext}
\end{axis}
\end{tikzpicture}
\caption{%
  Absolute forgetting curves on LoCoMo for the GPT-2~(124M) backbone at
  $1\times$ capacity.  Three methods (M.2, M.4, M.6) produce substantial
  retained-memory scores; three (M.1, M.3, M.5) are indistinguishable
  from the baseline.  Higher and flatter curves indicate stronger memory.%
}
\label{fig:forgetting_1x}
\end{figure}
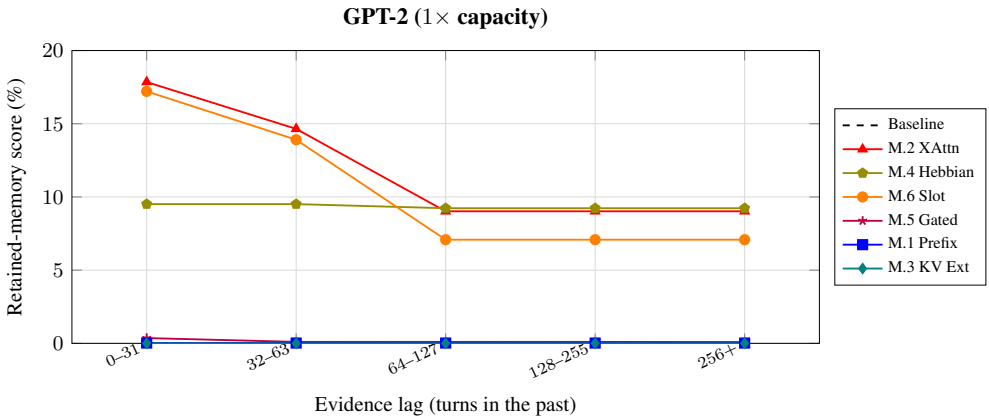

Figure~\ref{fig:forgetting_1x} shows the absolute forgetting curves for
all six methods on GPT-2 at $1\times$ capacity.  Three methods produce
substantial retained-memory scores: M.2~XAttn peaks at 17.8\% for
recent evidence and sustains 9.0\% beyond 256~turns; M.4~Hebbian
maintains a remarkably flat profile at 9.2--9.5\% across all lag bins;
M.6~Slot peaks at 17.2\% but decays to 7.1\% at longer lags.  The
remaining three methods---M.1~Prefix (0.02\%), M.3~KV~Extension (0.0\%),
and M.5~Gated (0.09--0.36\%)---produce scores indistinguishable from the
baseline.

\subsection{The inductive-bias dichotomy}
\label{sec:dichotomy}

The three successful methods share a common property: each injects
memory through a mechanism with strong inductive bias for integrating
external information into self-attention.

\begin{itemize}[leftmargin=*,itemsep=2pt]
\item \textbf{M.2~XAttn} inserts a full parallel cross-attention pathway
  (Flamingo-style), the most expressive injection mechanism.
\item \textbf{M.4~Hebbian} computes an associative recall
  $R_t = (H_t W_Q)\, M_{t-1}$ that explicitly routes content-addressed
  memory into new KV entries---a strong structural prior for
  pattern completion.
\item \textbf{M.6~Slot} uses sparse top-$k$ addressing to overwrite
  specific memory slots, giving the write rule a discrete selection
  mechanism that prevents dilution.
\end{itemize}

The three failing methods rely on weaker injection paths: M.1/M.3 use
shared or per-layer KV prefix projection (equivalent to adding soft
tokens), and M.5 filters through a content-dependent gate.  At
$1\times$ capacity these weaker priors cannot overcome the
distributional gap between pre-trained token representations and
persistent memory entries.

Table~\ref{tab:ranking} summarises the retained-memory scores and
knowledge gains for both capacity conditions.

\begin{table}[!hbtp]
\centering
\caption{Retained-memory score (min across lag bins, \%) and knowledge
accumulation ($\Delta K$) for each method at $1\times$ and $10\times$
capacity on GPT-2.  Bold: best in column.  Knowledge accumulation at
$10\times$ is omitted: the forgetting-curve convergence already
establishes the capacity-bridging result (\S\ref{sec:capacity}).}
\label{tab:ranking}
\small
\begin{tabular}{@{}l rr rr@{}}
\toprule
& \multicolumn{2}{c}{\textbf{$1\times$ capacity}}
& \multicolumn{2}{c}{\textbf{$10\times$ capacity}} \\
\cmidrule(lr){2-3}\cmidrule(lr){4-5}
\textbf{Method}
  & Ret.\ mem.\ (\%) & $\Delta K$
  & Ret.\ mem.\ (\%) & $\Delta K$ \\
\midrule
M.0 Baseline   &  0.00 & 5.57 &  0.00 & 5.57 \\
M.1 Prefix     &  0.02 & 0.00 &  9.20 & ---  \\
M.2 XAttn      &  9.02 & 7.34 &  9.88 & ---  \\
M.3 KV Ext     &  0.00 & 0.00 &  9.69 & ---  \\
M.4 Hebbian    &  \textbf{9.23} & 7.84 & \textbf{10.32} & ---  \\
M.5 Gated      &  0.09 & 0.17 &  7.62 & ---  \\
M.6 Slot       &  7.08 & \textbf{9.71} &  9.66 & ---  \\
\bottomrule
\end{tabular}
\end{table}

\subsection{Capacity bridging: $1\times$ vs.\ $10\times$}
\label{sec:capacity}

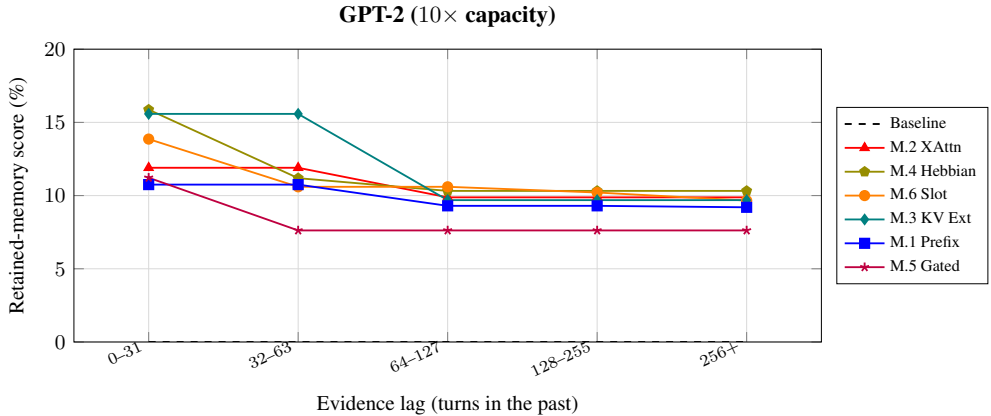
\begin{figure}[!hbtp]
\centering
\begin{tikzpicture}
\begin{axis}[
    width=0.92\linewidth,
    height=6cm,
    xlabel={Evidence lag (turns in the past)},
    ylabel={Retained-memory score (\%)},
    xmin=0.5, xmax=5.5,
    ymin=0, ymax=20,
    xtick={1,2,3,4,5},
    xticklabels={{$0$--$31$},{$32$--$63$},{$64$--$127$},{$128$--$255$},{$256+$}},
    x tick label style={rotate=25, anchor=east, font=\scriptsize},
    legend style={
      at={(1.02,0.5)},
      anchor=west,
      font=\scriptsize,
      cells={anchor=west},
    },
    grid=major,
    grid style={gray!30},
    every axis plot/.append style={thick},
    tick label style={font=\small},
    label style={font=\small},
    title={\textbf{GPT-2 ($10\times$ capacity)}},
]
\addplot[black, dashed] coordinates {(1,0) (2,0) (3,0) (4,0) (5,0)};
\addlegendentry{Baseline}
\addplot[red, mark=triangle*] coordinates {(1,11.90) (2,11.90) (3,9.88) (4,9.88) (5,9.88)};
\addlegendentry{M.2 XAttn}
\addplot[olive, mark=pentagon*] coordinates {(1,15.87) (2,11.19) (3,10.32) (4,10.32) (5,10.32)};
\addlegendentry{M.4 Hebbian}
\addplot[orange, mark=otimes*] coordinates {(1,13.86) (2,10.60) (3,10.60) (4,10.21) (5,9.66)};
\addlegendentry{M.6 Slot}
\addplot[teal, mark=diamond*] coordinates {(1,15.58) (2,15.58) (3,9.69) (4,9.69) (5,9.69)};
\addlegendentry{M.3 KV Ext}
\addplot[blue, mark=square*] coordinates {(1,10.75) (2,10.75) (3,9.30) (4,9.30) (5,9.20)};
\addlegendentry{M.1 Prefix}
\addplot[purple, mark=star] coordinates {(1,11.22) (2,7.62) (3,7.62) (4,7.62) (5,7.62)};
\addlegendentry{M.5 Gated}
\end{axis}
\end{tikzpicture}
\caption{%
  Absolute forgetting curves at $10\times$ capacity.  All six methods
  now produce substantial retained-memory scores.  Methods that failed
  at $1\times$ (M.1, M.3, M.5) converge to comparable performance with
  the strong-bias methods, demonstrating that the $1\times$ gap is
  architectural, not fundamental.%
}
\label{fig:forgetting_10x}
\end{figure}

Figure~\ref{fig:forgetting_10x} shows the forgetting curves at
$10\times$ capacity.  All six methods now produce substantial
retained-memory scores (7.6--10.3\%), and the three methods that
failed at $1\times$ match or approach the performance of the three
that succeeded.  M.1~Prefix jumps from 0.02\% to 9.20\%;
M.3~KV~Extension from 0.0\% to 9.69\%; M.5~Gated from 0.09\% to
7.62\%.

This convergence demonstrates that the $1\times$ failure is not a
fundamental incompatibility but a capacity bottleneck: with sufficient
memory dimensionality, even weak injection mechanisms can bridge the
distributional gap between pre-trained representations and persistent
memory entries.  The strong-bias methods (M.2, M.4, M.6) achieve this
with $10\times$ less capacity.

\subsection{Knowledge accumulation}
\label{sec:knowledge}

\begin{figure}[!hbtp]
\centering
\begin{tikzpicture}
\begin{axis}[
    width=0.92\linewidth,
    height=6cm,
    xlabel={Session},
    ylabel={Knowledge score $K_s$},
    xmin=0, xmax=29,
    ymin=0, ymax=14,
    xtick={0,5,10,15,20,25,29},
    legend style={
      at={(1.02,0.5)},
      anchor=west,
      font=\scriptsize,
      cells={anchor=west},
    },
    grid=major,
    grid style={gray!30},
    every axis plot/.append style={thick},
    tick label style={font=\small},
    label style={font=\small},
    title={\textbf{Knowledge accumulation ($1\times$)}},
]
\addplot[black, dashed] coordinates {
  (0,0) (1,0.28) (2,1.03) (3,2.82) (4,2.99) (5,3.80) (6,4.48)
  (7,4.42) (8,4.11) (9,4.90) (10,4.63) (11,4.63) (12,5.15)
  (13,5.02) (14,5.04) (15,5.84) (16,5.75) (17,6.64) (18,6.76)
  (19,6.82) (20,6.99) (21,6.62) (22,6.75) (23,7.32) (24,7.26)
  (25,5.81) (26,6.03) (27,5.89) (28,5.60) (29,5.57)
};
\addlegendentry{Baseline ($\Delta K\!=\!5.57$)}
\addplot[orange, mark=otimes*, mark repeat=3] coordinates {
  (0,0) (1,4.91) (2,5.48) (3,8.02) (4,9.84) (5,10.68) (6,10.81)
  (7,10.01) (8,9.25) (9,10.89) (10,10.12) (11,10.49) (12,9.11)
  (13,9.22) (14,8.66) (15,10.47) (16,10.50) (17,11.39) (18,11.36)
  (19,11.34) (20,11.30) (21,10.85) (22,11.17) (23,11.35) (24,10.38)
  (25,11.21) (26,11.64) (27,9.76) (28,10.34) (29,9.71)
};
\addlegendentry{M.6 Slot ($\Delta K\!=\!9.71$)}
\addplot[olive, mark=pentagon*, mark repeat=3] coordinates {
  (0,2.78) (1,4.04) (2,5.70) (3,9.22) (4,8.78) (5,7.24) (6,7.65)
  (7,12.07) (8,9.94) (9,10.94) (10,8.83) (11,10.92) (12,9.50)
  (13,9.52) (14,7.26) (15,11.27) (16,10.15) (17,11.93) (18,10.70)
  (19,9.82) (20,11.44) (21,9.52) (22,10.94) (23,11.81) (24,9.96)
  (25,10.73) (26,11.77) (27,8.85) (28,10.81) (29,10.62)
};
\addlegendentry{M.4 Hebbian ($\Delta K\!=\!7.84$)}
\addplot[red, mark=triangle*, mark repeat=3] coordinates {
  (0,3.70) (1,2.82) (2,4.97) (3,6.93) (4,8.31) (5,9.57) (6,10.03)
  (7,11.00) (8,10.42) (9,10.11) (10,8.92) (11,9.42) (12,8.80)
  (13,8.99) (14,9.82) (15,11.71) (16,11.59) (17,10.55) (18,10.32)
  (19,11.23) (20,10.98) (21,10.08) (22,10.08) (23,11.42) (24,10.76)
  (25,9.16) (26,10.41) (27,10.44) (28,10.11) (29,11.04)
};
\addlegendentry{M.2 XAttn ($\Delta K\!=\!7.34$)}
\end{axis}
\end{tikzpicture}
\caption{%
  Cumulative knowledge score $K_s$ over 30~sessions at $1\times$
  capacity.  Successful methods (M.2, M.4, M.6) accumulate knowledge
  substantially above the baseline; failing methods (M.1, M.3, M.5,
  not shown for clarity) produce $\Delta K < 0.2$.%
}
\label{fig:knowledge}
\end{figure}
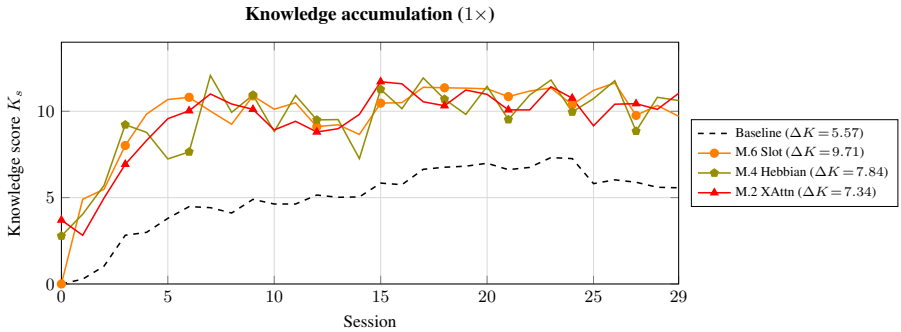

Figure~\ref{fig:knowledge} shows the cumulative knowledge score $K_s$
across 30~sessions at $1\times$ capacity.  The same dichotomy appears:
three methods accumulate knowledge well above the stateless baseline
($\Delta K = 5.57$), while three produce near-zero knowledge gain.
M.6~Slot achieves the highest $\Delta K = 9.71$, followed by
M.4~Hebbian ($\Delta K = 7.84$) and M.2~XAttn ($\Delta K = 7.34$).
M.1~Prefix and M.3~KV~Extension produce $\Delta K = 0.0$;
M.5~Gated reaches only $\Delta K = 0.17$.

Notably, the knowledge ranking differs from the forgetting-curve
ranking: M.6~Slot leads on $\Delta K$ despite having the lowest
retained-memory score among the three successful methods.  This suggests
that slot-based sparse writes are particularly well suited for
\emph{accumulating} diverse factual knowledge, even though the readout
fidelity per-fact is somewhat lower than for M.2 or M.4.

\section{Discussion}
\label{sec:discussion}

The results confirm that persistent latent-space memory transfers from
encoder--decoder to decoder-only models.  We discuss four aspects:
the architectural affinity hierarchy (\S\ref{sec:affinity}),
why the injection problem is harder yet the fit is more natural
(\S\ref{sec:harder}), why the write rule transfers without modification
(\S\ref{sec:universal_write}), and the current limitations of the study
(\S\ref{sec:limits}).

\subsection{Architectural affinity for persistent memory}
\label{sec:affinity}

We conjectured in \S\ref{sec:intro} that persistent memory has a natural
affinity ordering across transformer families:
decoder-only $>$ encoder--decoder $>$ encoder-only.
The reasoning rests on three properties.

\paragraph{Sequential processing.}
Autoregressive decoders process input one token at a time, building a
running KV cache that is conceptually a \emph{volatile} session memory.
Persistent memory generalises this cache across session boundaries---a
minimal conceptual leap.  Encoder--decoder models process context
bidirectionally in the encoder, then read it once in the decoder;
the temporal structure is less direct.  Encoder-only models
process everything in a single bidirectional pass with no notion of
incremental accumulation.

\paragraph{Dominant deployment.}
Virtually all modern deployed LLMs (GPT-4, Claude, LLaMA, Gemini,
Mistral) are decoder-only.  Because persistent memory retrofits a
frozen backbone, the practical impact is maximised when the paradigm
aligns with the most widely used architecture.

\paragraph{Generation capability.}
Persistent memory is useful only if the model can \emph{produce output}
conditioned on recalled information.  Decoder-only models generate
natively; encoder--decoder models generate through the decoder half;
encoder-only models require a separate head or decoder to produce text.
The more direct the generation path, the more cleanly memory recall
integrates into output.

\subsection{The injection paradox: harder mechanics, better fit}
\label{sec:harder}

The encoder--decoder setting benefits from cross-attention heads that
were pre-trained to integrate external representations.  Decoder-only
models must repurpose self-attention, which was trained to attend only to
preceding tokens in the same sequence.  The memory adapter must therefore
learn a harder mapping: projecting persistent state into a format that
self-attention treats as ``legitimate earlier context'' despite never
having seen such entries during pre-training.

This creates an apparent paradox: the architecture with the highest
conjectured affinity for persistent memory presents the hardest
\emph{injection} problem.  The fact that it works at all---and with comparable
retained-memory scores---suggests that self-attention heads possess
representational slack analogous to the cross-attention slack observed
in encoder--decoder models.  KV prefix injection exploits this
slack directly: the frozen self-attention computes
$\softmax(Q [K_{\mem}; K_{\text{input}}]^\top) [V_{\mem}; V_{\text{input}}]$,
treating memory entries as if they were earlier tokens in the sequence.

\subsection{The universal write rule}
\label{sec:universal_write}

The write rule $P_t = \gamma P_{t-1} + A_t^\top V_w$ is a
content-addressed aggregation over the current hidden state, gated by a
temporal decay factor.  It is architecture-agnostic because it depends
only on the existence of dense hidden states~$H_t \in \R^{n \times d}$,
which every transformer produces.  The read path must adapt to the
backbone's attention structure, but the write path does not.

This separation---universal write, architecture-specific read---is the
central structural insight.  It means that the memory bank~$P$ is a
general-purpose persistent substrate that can be connected to any frozen
backbone through a thin, architecture-specific adapter layer.

\subsection{Limitations}
\label{sec:limits}

All experiments use a single benchmark (LoCoMo) and a single training
seed.  Decoder-only evaluation reformulates LoCoMo's QA format for
autoregressive generation, which may introduce format-specific variance.
We do not test encoder-only models (BERT-style) or vision--language
models, which remain important future targets.  The write-side
projections are fixed random maps; learning them may improve performance
but would require backpropagation through the full conversation history.

\section{Conclusion}
\label{sec:conclusion}

We have demonstrated that persistent latent-space memory, originally
developed for frozen encoder--decoder models~\citep{jeong2026frozen},
transfers to the decoder-only setting despite the absence of
cross-attention injection points.  The central finding is an
inductive-bias dichotomy: at constrained capacity ($1\times$), only
memory methods with strong architectural priors---parallel
cross-attention (M.2), Hebbian associative recall (M.4), and slot-based
sparse write (M.6)---succeed, while the remaining three methods require
$10\times$ capacity to reach comparable performance.  This reveals that
the gap is one of efficiency rather than fundamental compatibility.

The architectural decomposition into a \emph{universal write rule}
(shared across backbones) and \emph{architecture-specific read paths}
(KV prefix, inserted cross-attention, gated branch) provides a
principled design template.  Because the write rule depends only on the
existence of dense hidden states, extending persistence to new
architecture families requires only a thin read adapter for the frozen
backbone.

Together with the encoder--decoder results
of~\citet{jeong2026frozen} and the brain-inspired memory modules
of~\citet{jeong2026lateral,jeong2026brain}, these findings establish
persistent latent-space memory as a general paradigm spanning the major
transformer families.  Extending the framework to encoder-only and
multimodal architectures, scaling to larger backbones, and learning
the write projections end-to-end remain promising directions for
future work.

\section*{Acknowledgements}
The author thanks Inha University in Tashkent for research support.
This work reflects the author's ongoing inquiry into nature and human cognition.

\section*{Reproducibility Statement}

All experiments use GPT-2 (124\,M parameters) from the Hugging Face
\texttt{transformers} library in bfloat16 precision.  Memory adapter
hyperparameters are reported in \S\ref{sec:training} and
Table~\ref{tab:capacity}.  The evaluation protocol follows
\citet{jeong2026frozen} without modification; the benchmark is the
publicly available LoCoMo dataset~\citep{maharana2024locomo}.  Training
uses a single NVIDIA GPU with AdamW ($\eta = 10^{-4}$).
Code and trained adapter weights will be released upon publication.

\section*{Ethics and Broader Impact}

Persistent memory enables language models to accumulate user-specific
information across sessions.  This capability raises privacy
considerations: remembered facts may include sensitive personal data.
Practical deployments should provide users with mechanisms to inspect,
edit, and delete stored memory entries, consistent with data-protection
regulations.  On the positive side, persistent memory reduces the need
for repeatedly restating context, which may improve accessibility for
users who struggle with extended text input.  Our experiments use a
publicly available benchmark and do not involve private user data.

\end{document}